\documentclass[fleqn,10pt]{wlscirep}
\usepackage[utf8]{inputenc}
\usepackage[T1]{fontenc}
\usepackage{soul}
\usepackage{lineno}

\title{A crowdsourced dataset of aerial images with annotated solar photovoltaic arrays and installation metadata\footnote{Revised preprint resubmitted to Scientific Data on 08/12/2022}}

\author[1,2*$\dag$]{Gabriel Kasmi}
\author[1$\dag$]{Yves-Marie Saint-Drenan}
\author[3$\dag$]{David Trebosc}
\author[1$\dag$]{Raphaël Jolivet}
\author[4]{Jonathan Leloux}
\author[4]{Babacar Sarr}
\author[2]{Laurent Dubus}

\affil[1]{Mines Paris - PSL University, Centre Observation, Impacts, Energy (O.I.E.), 06904 Sophia Antipolis, France}
\affil[2]{RTE France, 7C place du Dôme 92073 Paris La Défense, France}
\affil[3]{BDPV, 1 Rue du Capitaine Fracasse, 31320 Castanet Tolosan, France}
\affil[4]{LuciSun, Rue Saint-Jean, 29, Sart-Dames-Avelines, Belgium }

\affil[*]{corresponding author: Gabriel Kasmi (firstname.lastname@minesparis.psl.eu)}
\affil[$\dag$]{these authors contributed equally to this work}

\begin{abstract} 
Photovoltaic (PV) energy generation plays a crucial role in the energy transition. Small-scale, rooftop PV installations are deployed at an unprecedented pace, and their safe integration into the grid necessitates up-to-date, high-quality information. Overhead imagery is increasingly used to improve the knowledge of rooftop PV installations with machine learning models capable of automatically mapping these installations. However, these models cannot be reliably transferred from one region or imagery source to another without incurring a decrease in accuracy. To address this issue, known as distribution shift, and foster the development of PV array mapping pipelines, we propose a dataset containing aerial images, segmentation masks, and installation metadata. We provide installation metadata for more than 28000 installations. We supply ground truth segmentation masks for 13000 installations, including 7000 with annotations for two different image providers. Finally, we provide installation metadata that matches the annotation for more than 8000 installations. Dataset applications include end-to-end PV registry construction, robust PV installations mapping, and analysis of crowdsourced datasets. 
\end{abstract} 

\begin{document}

\flushbottom
\maketitle

\thispagestyle{empty}

\section*{Background \& Summary}

In 2021, photovoltaic (PV) power generation amounted to 821 $TWh$ worldwide and 14.3 $TWh$ in France\cite{rte2021bilan}. With an installed capacity of about 633 $GW_p$ worldwide\cite{iea2022solar} and 13.66 $GW_p$ in France, PV energy represents a growing share of the energy supply. 
The integration of growing amounts of solar energy in energy systems requires an accurate estimation of the produced power to maintain a constant balance between demand and supply. However, small-scale PV installations are generally invisible to transmission system operators (TSOs), meaning their generated power is not monitored\cite{shaker2015data}. For TSOs, the lack of reliable rooftop PV measurements increases the flexibility needs, i.e., the ability of the grid to compensate for load or supply variability\cite{kazmi2022good,saint2016analysis,saint2019bayesian,huber2014integration}. Estimating the PV power generation from meteorological data is common practice to overcome the lack of power measurements. However, this necessitates precise information on its installed capacity and metadata\cite{Saint-Drenan2017,Killinger2018}. Detailed information regarding small-scale PV is of interest for integrating renewable energies into the grid\cite{de2020monitoring}, or for understanding the factors behind its development\cite{wang2022deepsolar++}.

Currently, such PV installation registries covering large areas are neither easily available nor available everywhere. Recent research to construct global PV inventories\cite{dunnett2020harmonised,kruitwagen2021global} is limited to solar farms and does not include rooftop PV. A recent crowdsourcing effort enabled to map 86\% of rooftop and rooftop PV installations\cite{stowell2020harmonised}, but only in the United Kingdom. Other available datasets are aggregated at the communal scale (census level)\cite{yu2018deepsolar}.  

Remote sensing-based methods\cite{kruitwagen2021global,yu2018deepsolar,zech2020predicting,malof2016automatic} recently emerged as a promising solution to quickly and cheaply acquire detailed information on PV installations\cite{hu2022you}. These methods rely on overhead imagery and deep neural networks. The DeepSolar initiative led to the mapping of rooftop PV installations over the continental United-States\cite{yu2018deepsolar} or the state of North-Rhine Westphalia\cite{mayer20223d}. 
These remote sensing-based methods cannot scale to unseen regions without a sharp decrease in accuracy\cite{wang2017poor,kasmi2022towards}. It is caused by the sensitivity of deep learning models to distribution shifts\cite{torralba2011unbiased} (i.e., when {\it "the training distribution differs from the test distribution\cite{koh2021wilds}"}). These distribution shifts typically correspond to acquisition conditions and architectural differences across regions\cite{tuia2016domain}. The lack of robustness towards distribution shifts limits the reliability of deep learning-based registries for constructing official PV statistics\cite{de2020monitoring}. Therefore, developing PV mapping algorithms that are robust to distribution shifts is necessary. 

To encourage the development of such algorithms, we introduce a training dataset containing data for (i) addressing distribution shifts in remote sensing applications and (ii) helping design algorithms capable of extracting small-scale PV metadata from overhead imagery. 

To address distribution shifts, we gathered ground-truth annotations from two image providers for installations located in France. The double annotation enables researchers to evaluate the robustness of their approach to a shift in data provider (which affects acquisition conditions, acquisition device, and ground sampling distance) while keeping the same observed object. Our dataset provides ground truth installation masks for 13303 images from Google Earth\cite{gorelick2017google} and 7686 images from the French national institute of geographical and forestry information (IGN). To address architectural differences, researchers can either use the coarse-grained location included in our dataset or use our dataset in conjunction with other training datasets that mapped different areas (e.g., Bradbury {\it et al.}\cite{bradbury2016distributed} or Khomiakov {\it et al.} \cite{khomiakov2022solardk}). 

To extract PV systems' metadata, we release the installation metadata for 28807 installations. This metadata includes installed capacity, surface, tilt, and azimuth angles, sufficient for regional PV power estimation\cite{Saint-Drenan2017}. We linked the installation metadata and the ground truth images for 8019 installations. To the best of our knowledge, it is the first time a training dataset contains PV panel images, ground truth labels, and installations' metadata. We hope this data contributes to the ongoing effort to construct more detailed PV registries.

We obtained our labels through two crowdsourcing campaigns conducted in 2021 and 2022. Crowdsourcing is common practice in the machine learning community for annotating training datasets\cite{lin2014microsoft,deng2009imagenet}. We developed our crowdsourcing platform, and we were able to collect up to 50 annotations per image to maximize the accuracy of our annotations. Besides, multiple annotations per image facilitate measurement of the annotator's agreement or limit their individual annotations biases. Indeed, we found that some annotators were more cautious when annotating than others. We make the raw crowdsourcing data publicly available. It enables the replication of our annotations, but we also hope this will help research crowdsourcing, e.g., on the efficient combination of labels\cite{lefort2022improve}.

Our dataset targets practitioners and researchers in machine learning and crowdsourcing. Our data can serve as training data for remote PV mapping algorithms and test machine learning models' robustness against acquisition conditions shift. Additionally, we release the raw annotation data from the crowdsourcing campaigns for the community to carry out further studies on the fusion of multiple annotations into ground truth labels. The training dataset and the data coming from the crowdsourcing campaigns are accessible on our Zenodo repository\cite{kasmi2022bdappv}. 

\section*{Methods}

\begin{figure}[ht]
\centering
\includegraphics[width=0.9\linewidth]{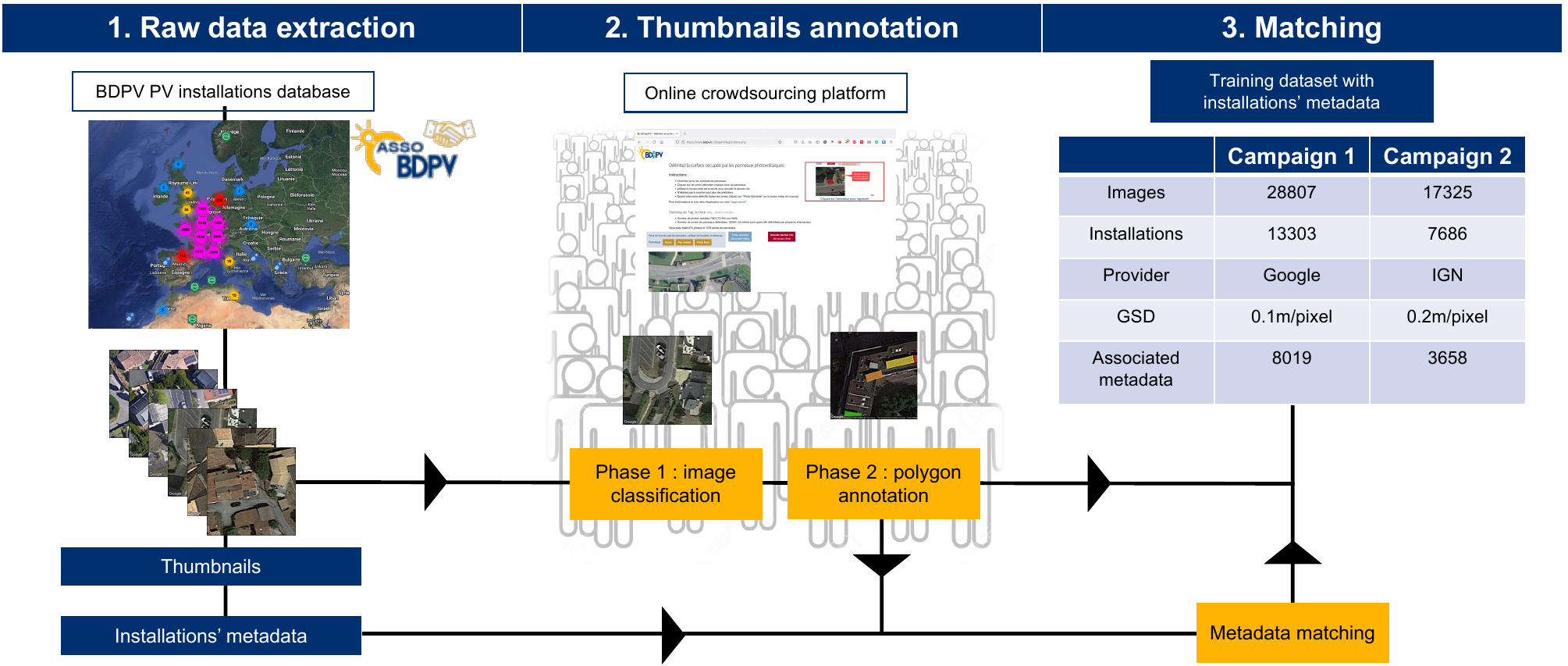}
\caption{Flowchart of the training dataset generation based on the BDPV PV data and crowdsourcing. "GSD" stands for the ground sampling distance, i.e., the distance between the centers of two adjacent pixels measured on the ground.}
\label{fig:flowchart}
\end{figure}

We illustrate our training dataset generation workflow in \autoref{fig:flowchart}. It comprises three main steps: thumbnails extraction, annotation of solar arrays, and metadata matching.

\subsection*{Thumbnails extraction}

Our annotation campaign leverages the database of PV systems operated by the non-profit association {\it Asso BDPV} ({\it Base de données Photovoltaïque} - Photovoltaic database). {\it Asso BDPV}  (BDPV) gathers metadata (geolocation and metadata of the PV systems) and the energy production data of PV installations provided by individual system owners, mainly in France and Western Europe. The primary purpose of the BDPV database is to monitor system owners' energy production. BDPV also promotes PV energy by disseminating information and data to the general public and public authorities. 

The BDPV data contains the localization of more than 28000 installations. We used this localization to extract the panels' thumbnails. During the first annotation campaigns, we extracted 28807 thumbnails using Google Earth Engine (GEE)\cite{gorelick2017google} application programming interface (API). For the second campaign, we extracted 17325 thumbnails from the IGN Geoservices portal (\url{https://geoservices.ign.fr/bdortho}).

Our thumbnails all have a resolution of 400$\times$400 pixels. Thumbnails extracted from GEE API correspond to a ground sampling distance (GSD) of 0.1 m/pixel. The API directly generates this thumbnail by setting the zoom level to 20, the localization to the ground truth localization contained in BDPV, and the output size to be 400$\times$400 pixels. For IGN images, the resolution of the thumbnails corresponds to a GSD of 0.2 m/pixel. The procedure for generating IGN thumbnails differs from Google. First, we downloaded geo-localized tiles from IGN's Geoservices portal. These tiles have a resolution of 25000$\times$25000 pixels, covering an area of 25 square kilometers. Then, extracted the thumbnail by generating a 400$\times$400 pixels raster centered around the location of the PV panel. Finally, we export this raster as a {\tt .png} file. We do not publish the exact location of the panels for confidentiality reasons.

The crowdsourcing campaigns took place on a dedicated platform called BDAPPV, which stands for {\it "Base de données apprentissage profond PV"} (PV database for deep learning). The BDAPPV platform is a web page where users can ergonomically annotate aerial images by clicking on the panel (phase 1) or delineating polygons around the PV panels (phase 2). \autoref{tab:records} summarizes the contribution of the annotators during the crowdsourcing campaigns. The web page is accessible at this URL: \url{https://www.bdpv.fr/_BDapPV/}.

\begin{table}[ht]
\centering
\begin{tabular}{|l|l|l|}
\hline
 & {\bf Google} & {\bf IGN} \\
\hline
Total number of actions & 349394 & 119528 \\
\hline
Total number of annotators & 1901 & 1021 \\
\hline
Actions during phase 1 &  291597 & 90084 \\
\hline
Actions during phase 2 & 68162 & 29444\\
\hline
Active annotators during phase 1 & 1043 & 51 \\
\hline
Active annotators during phase 2 & 960 & 980 \\
\hline
Active annotators during both phases & 102  & 10\\
\hline
\end{tabular}
\caption{\label{tab:records}Summary statistics of the contributions during the crowdsourcing campaigns.}
\end{table}

\subsection*{Annotation of solar arrays} 

We extracted thumbnails based on the geolocation of the installations recorded in the BDPV dataset. However, this geolocation can be inaccurate, so before asking users to draw polygons of PV installations, we asked them to classify the images. It corresponds to the first phase of the annotation campaign. Once users classified images, we asked users to draw the PV polygons on the remaining images. It corresponds to the second phase of the crowdsourcing campaign. 

We designed our campaign to get at least five annotations per image. It enabled us to derive metrics, which we term as {\it consensus metrics}, targeted at maximizing the quality of our labels. This way, we go further than the consensus between two annotators reported in previous work\cite{bradbury2016distributed} to measure annotation quality. The analysis of the users' annotations during phases 1 and 2 are reproducible using the notebook {\tt annotations} available on the public repository.

\subsubsection*{Phase 1: image classification}

During the first phase, the user clicks on an image if it depicts a PV panel. We recorded the localization of the user's click and instructed them to click {\it on} the PV panel if there was one. We collected an average of 10 actions (click with localization or no click) per image. The left panel of \autoref{fig:quality_phases_1_2} provides an example of annotations during phase 1. 
We apply the kernel density estimate (KDE) algorithm to the annotations to estimate a confidence level for the annotations and the approximate localization of the PV panel on the image. The likelihood $f_{\sigma}(x_i) $ of presence of a panel for each pixel $x_i$ is given by:

\begin{equation}
f_{\sigma}(x_i)=\frac{1}{N}\sum_{k=1}^{N}K_{\sigma}(x_k-x_i) 
\end{equation}

where $K_{\sigma}$ is a Gaussian kernel with a standard deviation $\sigma$, $x_k$ is the coordinate of the $k^{th}$ annotation, and $N$ is the total number of annotations. 

After an empirical investigation, we calibrated the standard deviation of the kernel to reflect the approximate spatial extent of an array on the image. We set its value to 25 pixels for Google images and 12 for IGN images. It corresponds to a distance of 2.5 m. As illustrated on \autoref{fig:quality_phases_1_2}, the KDE yields a heatmap whose hotspot locates on the solar array. The maximum value of the KDE quantifies the confidence level of the annotation. We refer to it as the {\it pixel annotation consensus} (PAC). This metric is proportional to the number of annotations. We use the PAC to determine whether an image contains an array. 

\subsubsection*{Phase 2: polygon annotation}

\begin{figure}[!h]
\centering
\includegraphics[width=0.9\linewidth]{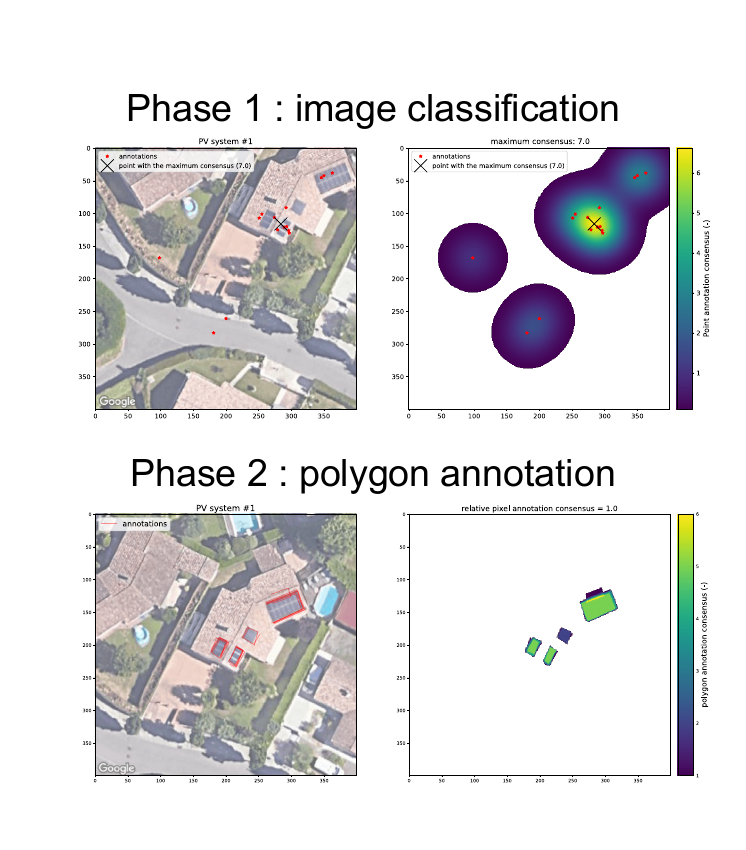}
\caption{Screenshot of the {\tt annotations} notebook, showing analysis of click annotations (phase 1, above) and polygon annotations (phase 2, below). During phase 1 (above), each red dot corresponds to an annotation. The density of annotations is greater near one of the panels, but we also see that other panels also received clicks.}
\label{fig:quality_phases_1_2}
\end{figure}

During the second phase, annotators delineate the PV panels on the images validated during phase 1. Users can draw as many polygons as they want. On average, we collected five polygons per image. We collect the coordinates of the polygons drawn by the annotators. As illustrated in the lower left panel of \autoref{fig:quality_phases_1_2}, a set of polygons is available for each array in an image. We can note from the annotation illustrated in \autoref{fig:quality_phases_1_2} that some polygons may be erroneous. However, these false positives have fewer annotations than true positives. To select only the true positives, we compute the PAC through the following steps:
\begin{enumerate}
    \item We convert each user's polygon into a binary raster;
    \item We compute the normalized PAC by summing all rasters and dividing by the number of annotators,
    \item We apply a relative threshold and keep only the pixels whose PAC is greater than the threshold;
    \item We compute the coordinates of the resulting mask using OpenCV's polygon detection algorithm (\url{https://docs.opencv.org/3.4/d4/d73/tutorial_py_contours_begin.html}).
\end{enumerate}
In step 2., the unnormalized PAC takes values between 0 and the number $N_i$ of annotators for the $i^{th}$ image. 0 means no user included the pixel into his polygon, and $N_i$ means that {\it all} annotators encapsulated the corresponding pixel in their polygons. 

\subsection*{Metadata matching}

Once we generate our segmentation masks, we match them with the installations' metadata reported in the BDPV dataset. Our matching procedure follows three steps: internal consistency, unique matching, and external consistency. Note that we only apply these filters when matching the metadata and the masks. 

\paragraph{Internal consistency} ensures that the entries in the BDPV dataset are coherent before any matching. It is simply a cleaning of the raw dataset. To do this cleaning, we verify whether the information in one column is coherent with the records from the other columns. For instance, if a system's record says it has ten modules and a surface of 3 squared meters, this would mean that each PV module has a surface of 0.3 squared meters, which is impossible (the smallest size being 1.7 squared meters). 

\paragraph{Unique matching} Our segmentation masks may depict more than one array. It occurs if, for instance, more than one panel is on the image shown to the annotators. In this case, we adopt a conservative view: if the segmentation mask depicts more than one panel, we cannot know which corresponds to the installation reported in the BDPV dataset. We do not match the segmentation mask with an installation in this case. 

\paragraph{External consistency}  After internal consistency filtering and unique matching, only segmentation masks depicting only one panel whose metadata is coherent remain. A final filtering step consists in making sure that the characteristics reported in the database match those that can be deduced from the segmentation mask. We assess the adequacy between the surface of the installation's mask and its true surface reported in the BDPV dataset by computing the ratio between them. We keep only installations whose ratio is equal to 1 (with a tolerance bandwidth of $\pm$ 25\%). We apply this bandwidth to accommodate the possible approximations in the segmentation mask. The reported surface excludes the inter-panel space and the distortions induced by the panel's projection on the image, as images are not perfectly orthorectified.

\newpage
\section*{Data Records}

The data records consist of two separate datasets, accessible on our Zenodo repository\cite{kasmi2022bdappv}, at this URL: \url{https://zenodo.org/record/7358126}.

\begin{enumerate}
    \item The {\it training dataset}: input images, segmentation masks, and PV installations' metadata, 
    \item The {\it crowdsourcing and replication data}: annotations from the users, for each image, provided in {\tt .json} format and the raw installations' metadata.
\end{enumerate}

Besides, the source code and notebooks used to generate the masks from the users' annotations are accessible on our public Git repository at this URL: \url{https://git.sophia.mines-paristech.fr/oie/bdappv}. This repository contains the source code used to generate the segmentation masks. It contains the notebooks {\tt annotations} and {\tt metadata}, which can be used to visualize the threshold analysis or the metadata matching procedure.

\subsection*{Training dataset}

The training dataset containing RGB images, ready-to-use segmentation masks of the two campaigns, and the file containing PV installations' metadata is accessible on our Zenodo repository. It is organized as follows:

\begin{itemize}
    \item {\tt bdappv/} Root data folder
    \begin{itemize}
        \item {\tt google} / {\tt ign}  One folder for each campaign
        \begin{itemize}
            \item {\tt img} Folder containing all the images presented to the annotators. This folder contains 28807 images for Google and 17325 for IGN. We provide all images as {\tt .png} files.
            \item {\tt mask} Folder containing all segmentation masks generated from the polygon annotations of the annotators. This folder contains 13303 masks for Google and 7686 for IGN. We provide all masks as {\tt .png} files.
        \end{itemize}
        \item {\tt metadata.csv} The {\tt .csv} file with the metadata of the installations. \autoref{tab:metadata.csv} describes the attributes of this table.
    \end{itemize}
\end{itemize}

\subsection*{Crowdsourcing and replication data}

The Git repository contains the raw crowdsourcing data and all the material necessary to re-generate our training dataset and technical validation. It is structured as follows: the {\tt raw} subfolder contains the raw annotation data from the two annotation campaigns and the raw PV installations' metadata. The {\tt replication} subfolder contains the compiled data used to generate our segmentation masks. The {\tt validation} subfolder contains the compiled data necessary to replicate the analyses presented in the technical validation section. 

\begin{itemize}
    \item {\tt data/} Root data folder
    \begin{itemize}
        \item {\tt raw/} Folder containing the raw crowdsourcing data and raw metadata;
        \begin{itemize}
            \item {\tt input-google.json}: Input data containing all information on images and raw annotators' contributions for both phases (clicks and polygons) during the first annotation campaign;
            \item {\tt input-ign.json}: Input data containing all information on images and raw annotators' contributions for both phases (clicks and polygons) during the second annotation campaign;
            \item {\tt raw-metadata.csv}: The file containing the PV systems' metadata extracted from the BDPV database before filtering. It can be used to replicate the association between the installations and the segmentation masks, as done in the notebook {\tt metadata}. \autoref{tab:raw-metadata.csv} describes the attributes of the {\tt raw-metadata.csv} table.
        \end{itemize}
        \item {\tt replication/} Folder containing the compiled data used to generate the segmentation masks;
        \begin{itemize}
            \item {\tt campaign-google} / {\tt campaign-ign}. One folder for each campaign
            \begin{itemize}
            \item {\tt click-analysis.json}: Output on the click analysis, compiling raw input into a few best-guess locations for the PV arrays. This dataset enables the replication of our annotations;
            \item {\tt polygon-analysis.json}: Output of polygon analysis, compiling raw input into a best-guess polygon for the PV arrays.
            \end{itemize}
        \end{itemize}
        \item {\tt validation/} Folder containing the compiled data used for technical validation.
        \begin{itemize}
            \item {\tt campaign-google} / {\tt campaign-ign}. One folder for each campaign
            \begin{itemize}
            \item {\tt click-analysis-thres=1.0.json}: Output of the click analysis with a lowered threshold to analyze the effect of the threshold on image classification, as done in the notebook {\tt annotations};
            \item {\tt polygon-analysis-thres=1.0.json}: Output of polygon analysis, with a lowered threshold to analyze the effect of the threshold on polygon annotation, as done in the notebook {\tt annotations}.
            \end{itemize}
              \item {\tt metadata.csv} the filtered installations' metadata.
        \end{itemize}
    \end{itemize}
\end{itemize}

\subsubsection*{Raw crowdsourcing data and raw installations' metadata}

The files {\tt input-google.json} and {\tt input-ign.json} provide the raw crowdsourcing data for both annotation campaigns. Both files are identically formatted. They contain all metadata for all images and the associated annotators' contributions for both phases (clicks and polygons). It enumerates all the clicks and polygons attached to the image. Coordinates are expressed in pixels relative to the upper-left corner of the image. We also provide information on the click or the polygon (annotator's ID, date, country). When an image contains no clicks, the field Clicks list is empty. If it does not contain polygons, the field Polygons list is empty. \autoref{tab:input.json} describes the attributes of the {\tt input} tables.

The raw metadata datasheet corresponds to the extraction of the BDPV database. This file contains all the installations' metadata. \autoref{tab:raw-metadata.csv} provides a list of the complete attributes. We use this file as input to associate the segmentation masks to the installations' metadata. 

\subsubsection*{Replication of the image classification and polygon annotation}

We compiled the files {\tt click-analysis.json} and {\tt polygon-analysis.json} from the raw inputs to classify the images and generate the segmentation masks, respectively. We provide these files to enable users to replicate our classification process and the generation of our masks.

The output of click analysis contains a list of detected PV installations' positions for each image. Each image contains at least one point, corresponding to the number of panels found in it. The {\tt score} variable summarizes the PAC associated with each point. By construction, the {\tt click-analysis.json} files only contain points with a PAC greater than 2.0 (see the technical validation section for more details on the threshold tuning). \autoref{tab:click-analysis.json} describes the attributes of the {\tt click-analysis.json} data file.

The output of polygon analysis contains a list of polygons as a compilation of all polygons annotated by annotators. It contains one or more polygons for each image corresponding to the PV arrays. Analogous to the {\tt click-analysis} files, this file summarizes the polygon annotations of the users. The variable {\tt score} records the relative PAC associated with each polygon. By construction, the {\tt polygon-analysis.json} files only contain polygons with a relative PAC greater than 0.45 (see the technical validation section for more details on the threshold tuning). \autoref{tab:polygon-analysis.json} describes the attributes of the {\tt polygon-analysis.json} data files. 

\subsubsection*{Validation data}

We compiled the files {\tt click-analysis-thres=1.0.json} and {\tt polygon-analysis-thres=1.0.json} to enable users to study how the thresholds chosen to generate our annotations affect the segmentation masks that we generate. The notebook {\tt annotations} enables us to carry out this study.

The {\tt metadata.csv} file corresponds to the output file of the notebook {\tt metadata}. We provide this file to enable users to replicate our analysis of the fit between the filtered installations and their segmentation masks.

\section*{Technical Validation}

Throughout the generation of the training dataset, we tested whether the threshold values chosen to classify the images, construct the polygon and associate the polygons to the installations' metadata yielded as few errors as possible. We base our approach on a consensus metric to classify images and construct the polygons, namely the pixel annotation consensus (PAC). Thus, we improve on Bradbury {\it et al.}\cite{bradbury2016distributed}, who proposed a confidence value based on the Jaccard Similarity Index\cite{levandowsky1971distance} between the two annotations. As for the association between the polygons and installations' metadata, we balance between accuracy and keeping as many installations as possible.

\subsection*{Analysis of the consensus value for image classification}

As mentioned in the methods section, the choice criterion for image classification during phase 1 is the consensus among users. We empirically investigated a range of thresholds and determined that a value of 2.0 yielded the most accurate classification results. In other words, we require that at least three annotators click around the same point to validate the classification. 

We use an {\it absolute} (unnormalized by the number of annotators for this image) threshold to decide whether the image contains a panel. The threshold is absolute because users could only click once on the image during the annotation campaign, even if the latter contained more than one array. As such, an absolute threshold does not dilute the consensus among users when there is more than one panel on the image. 

The leftmost plot of \autoref{fig:validation} plots the histogram of the absolute PAC. Visual inspection revealed that the peak for values below 2.0 corresponded to false positives. We enable replication of the threshold analysis in the notebook {\tt annotation}. 

\subsection*{Analysis of the consensus value of the polygon annotation}

Like the click annotation, we used a consensus metric to merge the users' annotations. After empirical investigations, we found that a {\it relative} threshold (expressed as a share of the total number of annotators) was the most effective for yielding the most accurate masks and that its value should be 0.45. In other words, we consider that a pixel depicts an installation if at least 45\% of the annotators included it in their polygons.

The center plot of \autoref{fig:validation} depicts the histogram of the relative PAC. Visual inspection revealed that the few values below 0.45 corresponded to remaining false positives (e.g., roof windows). The use of a relative threshold is motivated by the fact that the users can annotate as many polygons as they want. We enable replication of the threshold analysis in the notebook {\tt annotation}. 

\subsection*{Consistency between annotations and metadata of the PV installations}

We link segmentation masks and annotation metadata according to the steps described in the section "Metadata matching ." To measure the quality of this linkage, we measure the Pearson correlation coefficient (PCC) between the surface reported in the installation' metadata dataset (referred to as the "target" surface) and the surface estimated from the segmentation masks (referred to as the "estimated" surface). The higher the PCC, the better our matching procedure. 

\begin{figure}[ht]
\centering
\includegraphics[width=\linewidth]{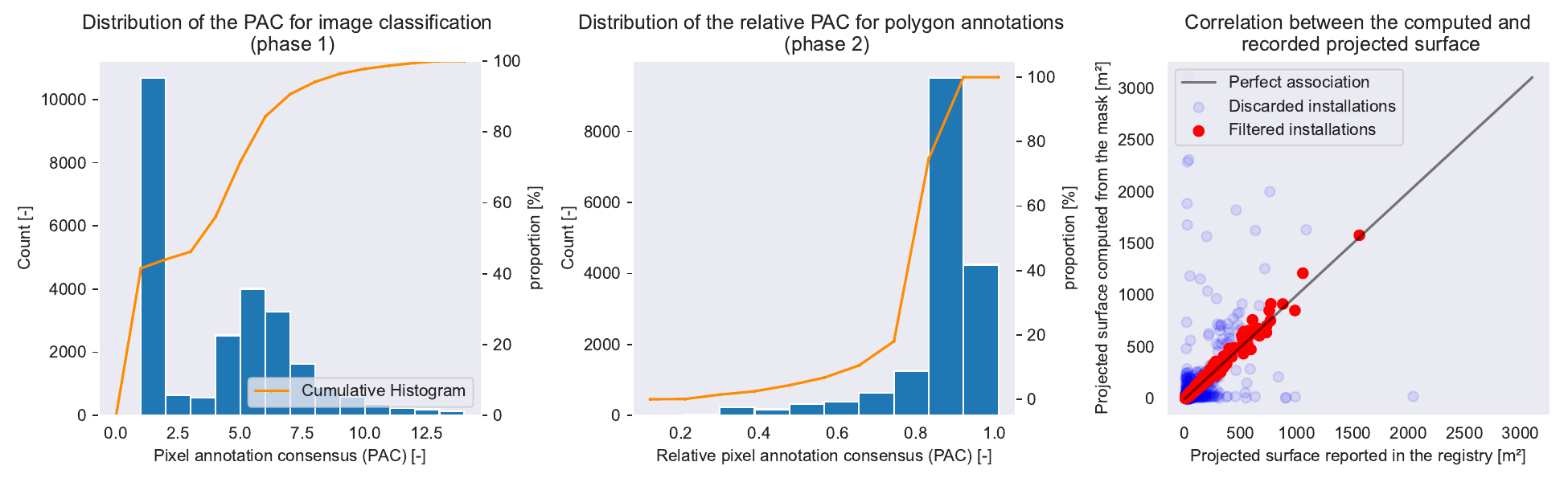}
\caption{Validation by comparison of the surface estimated from the masks and the surface reported in the PV installations' metadata.}
\label{fig:validation}
\end{figure}

\autoref{fig:validation} plots estimated and target surfaces. After filtering, we obtain a PCC coefficient of 0.99 between the target and estimated surfaces. Without filtering, the PCC coefficient equals 0.68 for Google images and 0.61 for IGN images. It shows that our metadata-matching procedure enabled us to pick the installations with the best fit between the observable metadata and masks. 

Our matching procedure comprises three steps: internal consistency, unicity and external consistency. Each of these steps discards installations from the BDPV database. \autoref{tab:filtering} summarizes the number of installations filtered at each process step. We can see that most of the filtering happens when we discard segmentation masks on which there is more than one installation.  

\begin{table}[ht]
    \centering
    \begin{tabular}{| l | l | l | l | l |}
    \hline
                    &  {\bf Google} & {\bf IGN} & {\bf Removed Google (\%)} & {\bf Removed IGN (\%)}\\
\hline
Raw                 & 28408 & 28408 &  0 (0) & 0 (0)\\
\hline
Internal consistency & 27780 & 27780 & 628 (2.21) & 628 (2.21)  \\
\hline
Mask uniqueness & 10523 & 5883 & 17257 (62.12) & 21897 (78.82)\\
\hline
External consistency & 8019 & 3658 & 2504 (23.80) & 2225 (37.82)\\
\hline
   
\end{tabular}
\caption{Number of installations filtered through the different filtering steps during the association between the masks and the installations' metadata.}
\label{tab:filtering}
\end{table}

\section*{Usage Notes}

We designed the complete dataset records to be directly used as training data in machine learning projects. The ready-to-use data is accessible on our Zenodo repository accessible at this URL \url{https://zenodo.org/record/7358126}. This repository also stores the raw crowdsourcing data and the files necessary to reproduce our segmentation masks and analyses. We compiled the files {\tt click-analysis.json} and {\tt polygon-analysis.json} using the Python scripts {\tt click-analysis.py} and {\tt polygon-analysis.py}, provided in our repository from the raw input data. This repository also contains the notebooks {\tt annotations} and {\tt metadata}. The notebook {\tt annotations} presents the analysis of crowdsourced data from the crowdsourcing campaigns. The notebook {\tt metadata} filters the {\tt raw-metadata.csv} datasheet.

Between phases 1 and 2, we generated new thumbnails re-centered on the PV installations. The new center corresponds to the coordinates of the estimated center of the (first in the list) detected PV installation. Therefore, to replicate the click analysis on the corresponding image, interested users need to download the corresponding image accessible on the BDAPPV website as illustrated in the notebook {\tt annotations}. We re-center images by generating a new thumbnail centered around an updated location, according to the procedure described in the section "methods."

The centering of the images will not induce a bias during learning because our thumbnails have a larger resolution (400 $\times$ 400) than the typical input size of typical neural networks (224 $\times$ 224). Adding a random crop transform during training will result in panels not being centered anymore. Besides, during the IGN campaign, we only re-centered about 13\% of the images. 

\section*{Code availability}

Our public repository accessible at this URL \url{https://git.sophia.mines-paristech.fr/oie/bdappv} contains the code to generate the masks, filter the metadata and analyze our results. Interested users can clone this repository to replicate our results or conduct analyses. 

\section*{Rights and permissions}

Code, raw crowdsourcing data, and compiled data are accessible on the project repository. All materials are provided under the CC-BY license. This license allows reusers to distribute, remix, adapt, and build upon the material in any medium or format, as long as attribution is given to the creator. The license allows for commercial use.

 This article is licensed under a Creative Commons Attribution 4.0 International License, which permits use, sharing, adaptation, distribution, and reproduction in any medium or format, as long as you give appropriate credit to the original author(s) and the source, provide a link to the Creative Commons license, and indicate if changes were made. The images or third-party material in this article are included in the article's Creative Commons license unless indicated otherwise in a credit line to the material. If material is not included in the article's Creative Commons license and your intended use is not permitted by statutory regulation or exceeds the permitted use. In that case, you will need to obtain permission directly from the copyright holder. To view a copy of this license, visit \url{http://creativecommons.org/licenses/by/4.0/}.

The Creative Commons Public Domain Dedication waiver \url{http://creativecommons.org/publicdomain/zero/1.0/} applies to the metadata files associated with this article.

\bibliography{sample}

\begin{thebibliography}{10}
\urlstyle{rm}
\expandafter\ifx\csname url\endcsname\relax
  \def\url#1{\texttt{#1}}\fi
\expandafter\ifx\csname urlprefix\endcsname\relax\def\urlprefix{URL }\fi
\expandafter\ifx\csname doiprefix\endcsname\relax\def\doiprefix{DOI: }\fi
\providecommand{\bibinfo}[2]{#2}
\providecommand{\eprint}[2][]{\url{#2}}

\bibitem{rte2021bilan}
\bibinfo{author}{{RTE France}}.
\newblock \bibinfo{title}{{Bilan électrique 2021}} (\bibinfo{year}{2022}).
\newblock \bibinfo{note}{\url{https://bilan-electrique-2021.rte-france.com/}}.

\bibitem{iea2022solar}
\bibinfo{author}{IEA}.
\newblock \bibinfo{title}{{Solar PV}} (\bibinfo{year}{2022}).
\newblock \bibinfo{note}{\url{https://www.iea.org/reports/solar-pv}}.

\bibitem{shaker2015data}
\bibinfo{author}{Shaker, H.}, \bibinfo{author}{Zareipour, H.} \&
  \bibinfo{author}{Wood, D.}
\newblock \bibinfo{journal}{\bibinfo{title}{{A data-driven approach for
  estimating the power generation of invisible solar sites}}}.
\newblock {\emph{\JournalTitle{IEEE Transactions on Smart Grid}}}
  \textbf{\bibinfo{volume}{7}}, \bibinfo{pages}{2466--2476}
  (\bibinfo{year}{2015}).

\bibitem{kazmi2022good}
\bibinfo{author}{Kazmi, H.} \& \bibinfo{author}{Tao, Z.}
\newblock \bibinfo{journal}{\bibinfo{title}{{How good are TSO load and
  renewable generation forecasts: Learning curves, challenges, and the road
  ahead}}}.
\newblock {\emph{\JournalTitle{Applied Energy}}}
  \textbf{\bibinfo{volume}{323}}, \bibinfo{pages}{119565}
  (\bibinfo{year}{2022}).

\bibitem{saint2016analysis}
\bibinfo{author}{Saint-Drenan, Y.-M.}, \bibinfo{author}{Good, G.~H.},
  \bibinfo{author}{Braun, M.} \& \bibinfo{author}{Freisinger, T.}
\newblock \bibinfo{journal}{\bibinfo{title}{{Analysis of the uncertainty in the
  estimates of regional PV power generation evaluated with the upscaling
  method}}}.
\newblock {\emph{\JournalTitle{Solar Energy}}} \textbf{\bibinfo{volume}{135}},
  \bibinfo{pages}{536--550} (\bibinfo{year}{2016}).

\bibitem{saint2019bayesian}
\bibinfo{author}{Saint-Drenan, Y.-M.} \emph{et~al.}
\newblock \bibinfo{journal}{\bibinfo{title}{{Bayesian parameterisation of a
  regional photovoltaic model--Application to forecasting}}}.
\newblock {\emph{\JournalTitle{Solar Energy}}} \textbf{\bibinfo{volume}{188}},
  \bibinfo{pages}{760--774} (\bibinfo{year}{2019}).

\bibitem{huber2014integration}
\bibinfo{author}{Huber, M.}, \bibinfo{author}{Dimkova, D.} \&
  \bibinfo{author}{Hamacher, T.}
\newblock \bibinfo{journal}{\bibinfo{title}{{Integration of wind and solar
  power in Europe: Assessment of flexibility requirements}}}.
\newblock {\emph{\JournalTitle{Energy}}} \textbf{\bibinfo{volume}{69}},
  \bibinfo{pages}{236--246} (\bibinfo{year}{2014}).

\bibitem{Saint-Drenan2017}
\bibinfo{author}{Saint-Drenan, Y.~M.}, \bibinfo{author}{Good, G.~H.} \&
  \bibinfo{author}{Braun, M.}
\newblock \bibinfo{journal}{\bibinfo{title}{{A probabilistic approach to the
  estimation of regional photovoltaic power production}}}.
\newblock {\emph{\JournalTitle{Solar Energy}}}
  \url{10.1016/j.solener.2017.03.007} (\bibinfo{year}{2017}).

\bibitem{Killinger2018}
\bibinfo{author}{Killinger, S.} \emph{et~al.}
\newblock \bibinfo{journal}{\bibinfo{title}{{On the search for representative
  characteristics of PV systems: Data collection and analysis of PV system
  azimuth, tilt, capacity, yield and shading}}}.
\newblock {\emph{\JournalTitle{Solar Energy}}} \textbf{\bibinfo{volume}{173}},
  \url{10.1016/j.solener.2018.08.051} (\bibinfo{year}{2018}).

\bibitem{de2020monitoring}
\bibinfo{author}{De~Jong, T.} \emph{et~al.}
\newblock \bibinfo{journal}{\bibinfo{title}{{Monitoring Spatial Sustainable
  Development: semi-automated analysis of Satellite and Aerial Images for
  Energy Transition and Sustainability Indicators}}}.
\newblock {\emph{\JournalTitle{arXiv preprint arXiv:2009.05738}}}
  (\bibinfo{year}{2020}).

\bibitem{wang2022deepsolar++}
\bibinfo{author}{Wang, Z.}, \bibinfo{author}{Arlt, M.-L.},
  \bibinfo{author}{Zanocco, C.}, \bibinfo{author}{Majumdar, A.} \&
  \bibinfo{author}{Rajagopal, R.}
\newblock \bibinfo{journal}{\bibinfo{title}{{DeepSolar++: Understanding
  residential solar adoption trajectories with computer vision and technology
  diffusion models}}}.
\newblock {\emph{\JournalTitle{Joule}}} \textbf{\bibinfo{volume}{6}},
  \bibinfo{pages}{2611--2625} (\bibinfo{year}{2022}).

\bibitem{dunnett2020harmonised}
\bibinfo{author}{Dunnett, S.}, \bibinfo{author}{Sorichetta, A.},
  \bibinfo{author}{Taylor, G.} \& \bibinfo{author}{Eigenbrod, F.}
\newblock \bibinfo{journal}{\bibinfo{title}{Harmonised global datasets of wind
  and solar farm locations and power}}.
\newblock {\emph{\JournalTitle{Scientific data}}} \textbf{\bibinfo{volume}{7}},
  \bibinfo{pages}{1--12} (\bibinfo{year}{2020}).

\bibitem{kruitwagen2021global}
\bibinfo{author}{Kruitwagen, L.} \emph{et~al.}
\newblock \bibinfo{journal}{\bibinfo{title}{A global inventory of photovoltaic
  solar energy generating units}}.
\newblock {\emph{\JournalTitle{Nature}}} \textbf{\bibinfo{volume}{598}},
  \bibinfo{pages}{604--610} (\bibinfo{year}{2021}).

\bibitem{stowell2020harmonised}
\bibinfo{author}{Stowell, D.} \emph{et~al.}
\newblock \bibinfo{journal}{\bibinfo{title}{{A harmonised, high-coverage, open
  dataset of solar photovoltaic installations in the UK}}}.
\newblock {\emph{\JournalTitle{Scientific Data}}} \textbf{\bibinfo{volume}{7}},
  \bibinfo{pages}{1--15} (\bibinfo{year}{2020}).

\bibitem{yu2018deepsolar}
\bibinfo{author}{Yu, J.}, \bibinfo{author}{Wang, Z.},
  \bibinfo{author}{Majumdar, A.} \& \bibinfo{author}{Rajagopal, R.}
\newblock \bibinfo{journal}{\bibinfo{title}{{DeepSolar: A machine learning
  framework to efficiently construct a solar deployment database in the United
  States}}}.
\newblock {\emph{\JournalTitle{Joule}}} \textbf{\bibinfo{volume}{2}},
  \bibinfo{pages}{2605--2617} (\bibinfo{year}{2018}).

\bibitem{zech2020predicting}
\bibinfo{author}{Zech, M.} \& \bibinfo{author}{Ranalli, J.}
\newblock \bibinfo{title}{{Predicting PV Areas in Aerial Images with Deep
  Learning}}.
\newblock In \emph{\bibinfo{booktitle}{2020 47th IEEE Photovoltaic Specialists
  Conference (PVSC)}}, \bibinfo{pages}{0767--0774}
  (\bibinfo{organization}{IEEE}, \bibinfo{year}{2020}).

\bibitem{malof2016automatic}
\bibinfo{author}{Malof, J.~M.}, \bibinfo{author}{Bradbury, K.},
  \bibinfo{author}{Collins, L.~M.} \& \bibinfo{author}{Newell, R.~G.}
\newblock \bibinfo{journal}{\bibinfo{title}{Automatic detection of solar
  photovoltaic arrays in high resolution aerial imagery}}.
\newblock {\emph{\JournalTitle{Applied energy}}}
  \textbf{\bibinfo{volume}{183}}, \bibinfo{pages}{229--240}
  (\bibinfo{year}{2016}).

\bibitem{hu2022you}
\bibinfo{author}{Hu, W.} \emph{et~al.}
\newblock \bibinfo{journal}{\bibinfo{title}{{What you get is not always what
  you see—pitfalls in solar array assessment using overhead imagery}}}.
\newblock {\emph{\JournalTitle{Applied Energy}}}
  \textbf{\bibinfo{volume}{327}}, \bibinfo{pages}{120143}
  (\bibinfo{year}{2022}).

\bibitem{mayer20223d}
\bibinfo{author}{Mayer, K.} \emph{et~al.}
\newblock \bibinfo{journal}{\bibinfo{title}{{3D-PV-Locator: Large-scale
  detection of rooftop-mounted photovoltaic systems in 3D}}}.
\newblock {\emph{\JournalTitle{Applied Energy}}}
  \textbf{\bibinfo{volume}{310}}, \bibinfo{pages}{118469}
  (\bibinfo{year}{2022}).

\bibitem{wang2017poor}
\bibinfo{author}{Wang, R.}, \bibinfo{author}{Camilo, J.},
  \bibinfo{author}{Collins, L.~M.}, \bibinfo{author}{Bradbury, K.} \&
  \bibinfo{author}{Malof, J.~M.}
\newblock \bibinfo{title}{The poor generalization of deep convolutional
  networks to aerial imagery from new geographic locations: an empirical study
  with solar array detection}.
\newblock In \emph{\bibinfo{booktitle}{2017 IEEE Applied Imagery Pattern
  Recognition Workshop (AIPR)}}, \bibinfo{pages}{1--8}
  (\bibinfo{organization}{IEEE}, \bibinfo{year}{2017}).

\bibitem{kasmi2022towards}
\bibinfo{author}{Kasmi, G.}, \bibinfo{author}{Dubus, L.},
  \bibinfo{author}{Blanc, P.} \& \bibinfo{author}{Saint-Drenan, Y.-M.}
\newblock \bibinfo{title}{{Towards unsupervised assessment with open-source
  data of the accuracy of deep learning-based distributed PV mapping}}.
\newblock In \emph{\bibinfo{booktitle}{Workshop on Machine Learning for Earth
  Observation (MACLEAN), in Conjunction with the ECML/PKDD 2022}}
  (\bibinfo{year}{2022}).

\bibitem{torralba2011unbiased}
\bibinfo{author}{Torralba, A.} \& \bibinfo{author}{Efros, A.~A.}
\newblock \bibinfo{title}{Unbiased look at dataset bias}.
\newblock In \emph{\bibinfo{booktitle}{CVPR 2011}}, \bibinfo{pages}{1521--1528}
  (\bibinfo{organization}{IEEE}, \bibinfo{year}{2011}).

\bibitem{koh2021wilds}
\bibinfo{author}{Koh, P.~W.} \emph{et~al.}
\newblock \bibinfo{title}{Wilds: A benchmark of in-the-wild distribution
  shifts}.
\newblock In \emph{\bibinfo{booktitle}{International Conference on Machine
  Learning}}, \bibinfo{pages}{5637--5664} (\bibinfo{organization}{PMLR},
  \bibinfo{year}{2021}).

\bibitem{tuia2016domain}
\bibinfo{author}{Tuia, D.}, \bibinfo{author}{Persello, C.} \&
  \bibinfo{author}{Bruzzone, L.}
\newblock \bibinfo{journal}{\bibinfo{title}{{Domain adaptation for the
  classification of remote sensing data: An overview of recent advances}}}.
\newblock {\emph{\JournalTitle{IEEE geoscience and remote sensing magazine}}}
  \textbf{\bibinfo{volume}{4}}, \bibinfo{pages}{41--57} (\bibinfo{year}{2016}).

\bibitem{gorelick2017google}
\bibinfo{author}{Gorelick, N.} \emph{et~al.}
\newblock \bibinfo{journal}{\bibinfo{title}{{Google Earth Engine:
  Planetary-scale geospatial analysis for everyone}}}.
\newblock {\emph{\JournalTitle{Remote sensing of Environment}}}
  \textbf{\bibinfo{volume}{202}}, \bibinfo{pages}{18--27}
  (\bibinfo{year}{2017}).

\bibitem{bradbury2016distributed}
\bibinfo{author}{Bradbury, K.} \emph{et~al.}
\newblock \bibinfo{journal}{\bibinfo{title}{{Distributed solar photovoltaic
  array location and extent dataset for remote sensing object
  identification}}}.
\newblock {\emph{\JournalTitle{Scientific data}}} \textbf{\bibinfo{volume}{3}},
  \bibinfo{pages}{1--9} (\bibinfo{year}{2016}).

\bibitem{khomiakov2022solardk}
\bibinfo{author}{Khomiakov, M.~M.} \emph{et~al.}
\newblock \bibinfo{title}{{SolarDK: A high-resolution urban solar panel image
  classification and localization dataset}}.
\newblock In \emph{\bibinfo{booktitle}{NeurIPS 2022 Workshop on Tackling
  Climate Change with Machine Learning}} (\bibinfo{year}{2022}).

\bibitem{lin2014microsoft}
\bibinfo{author}{Lin, T.-Y.} \emph{et~al.}
\newblock \bibinfo{title}{{Microsoft coco: Common objects in context}}.
\newblock In \emph{\bibinfo{booktitle}{European conference on computer
  vision}}, \bibinfo{pages}{740--755} (\bibinfo{organization}{Springer},
  \bibinfo{year}{2014}).

\bibitem{deng2009imagenet}
\bibinfo{author}{Deng, J.} \emph{et~al.}
\newblock \bibinfo{title}{{Imagenet: A large-scale hierarchical image
  database}}.
\newblock In \emph{\bibinfo{booktitle}{2009 IEEE conference on computer vision
  and pattern recognition}}, \bibinfo{pages}{248--255}
  (\bibinfo{organization}{Ieee}, \bibinfo{year}{2009}).

\bibitem{lefort2022improve}
\bibinfo{author}{Lefort, T.}, \bibinfo{author}{Charlier, B.},
  \bibinfo{author}{Joly, A.} \& \bibinfo{author}{Salmon, J.}
\newblock \bibinfo{journal}{\bibinfo{title}{{Improve learning combining
  crowdsourced labels by weighting Areas Under the Margin}}}.
\newblock {\emph{\JournalTitle{arXiv preprint arXiv:2209.15380}}}
  (\bibinfo{year}{2022}).

\bibitem{kasmi2022bdappv}
\bibinfo{author}{Kasmi, G.} \emph{et~al.}
\newblock \bibinfo{title}{{A crowdsourced dataset of aerial images with
  annotated solar photovoltaic arrays and installation metadata}},
  \url{10.5281/zenodo.7358126} (\bibinfo{year}{2022}).

\bibitem{levandowsky1971distance}
\bibinfo{author}{Levandowsky, M.} \& \bibinfo{author}{Winter, D.}
\newblock \bibinfo{journal}{\bibinfo{title}{Distance between sets}}.
\newblock {\emph{\JournalTitle{Nature}}} \textbf{\bibinfo{volume}{234}},
  \bibinfo{pages}{34--35} (\bibinfo{year}{1971}).

\end{thebibliography}

\section*{Acknowledgements} 

We want to thank all annotators who participated in the crowdsourcing campaigns. We also would like to thank the association BDPV and the users of the online forum {\it Forum Photovoltaïque} (\url{https://forum-photovoltaique.fr/}). They actively participated in the execution of the crowdsourcing elaboration of this project. 

This project is carried out as part of the Ph.D. thesis of Gabriel Kasmi, sponsored by the French transmission system operator RTE France and partly funded by the national agency for research and technology (ANRT) under the CIFRE contract 2020/0685.

The European Commission partially funds the work of Jonathan Leloux and Babacar Sarr through the Horizon 2020 project SERENDI-PV (\url{https://serendipv.eu/}), which belongs to the Research and Innovation Programme, under Grant Agreement 953016.

\section*{Author contributions statement}

G.K.: Contribution to the concept development, contributions to the manuscript review and editing of the final draft, second campaign management (generation of the thumbnails from IGN data for the two phases), matching between the metadata and the installations, \autoref{fig:flowchart}, \autoref{fig:validation} and tables formatting, crowdsourcing campaign analysis, manuscript revision. R.J.: Contribution to manuscript and code for annotations analysis (scripts and notebooks), generation of the ground truth labels, \autoref{fig:quality_phases_1_2} and tables formatting. Y.-M. S.-D.: development of the concept of BDAPPV, contribution to the manuscript, review, and editing of the final draft, analysis of the annotation, \autoref{fig:quality_phases_1_2} and \autoref{fig:validation} formatting, manuscript revision. D.T.: development of the concept of BDAPPV, conception and development of the annotation platform, communication with the community of annotators, collection and management of the annotation data, and contribution to the manuscript. J.L.: contribution to the concept development and review of the manuscript. B.S.: contribution to the concept development and review of the manuscript. L.D.:contribution to the concept development and review of the manuscript.

\section*{Competing interests} 

Y.-M. Saint-Drenan and D. Trebosc are members of the non-profit association Asso BDPV, which collects and maintains a database of rooftop PV plants. Y.-M. Saint-Drenan, D. Trebosc, J. Leloux, and B. Sarr are members of the initiative coBDPV, aiming to improve the analysis conducted within the BDPV platform. G. Kasmi is carrying out a Ph.D. funded by RTE (the French transmission system operator) and the national agency for research and technology (ANRT). L. Dubus is a senior research scientist at RTE. R. Jolivet declares no conflict of interest.

\section*{Tables}

\begin{table}[ht]
\centering
\begin{tabular}{|l|l|l|l|l|}
\hline
{\bf Field} & {\bf Attribute name} & {\bf Description} & {\bf Format} & {\bf Unit} \\
\hline
Image ID & id	&ID of image & String & -\\
\hline
City & city	&City of the image & String & - \\
\hline
Department & department&	Departement of the image & String & -\\
\hline
Region & region	&Region of the image\\
\hline
Installation ID &install\_id&	ID of the corresponding installation in & Integer & - \\
&& the BDPV database. & &  \\
\hline
Clicks list & clicks[]	&List of clicks & List & -\\
\hline
Click Pixel x coordinate & clicks[].x &	x position of click in image & Integer & Pixel\\
\hline
Click Pixel y coordinate & clicks[].y &	y position of click in image & Integer & Pixel\\
\hline
Click metadata & clicks[].action&	Metadata of the click action & List & -\\
\hline
Click country & clicks[].action.country&	Country of actor of the click & String & -\\
\hline
Click region &clicks[].action.region&	Region of actor of the click & String & - \\
\hline
Click date &clicks[].action.date &	Date / time of click & String & Date\\
\hline
Click author & clicks[].action.actorId &	ID of the author of the click & Integer & -\\
\hline
Polygons list & polygons[]	&List of polygons & List & -\\
\hline
Polygon points list &polygons[].points[]&	List of point of the polygon & List & -\\
\hline
Polygon point Pixel x coordinate & polygons[].points[].x	&x position of the point of polygon & Integer & Pixel\\
\hline
Polygon point Pixel y coordinate & polygons[].points[].y	&y position of the point of polygon & Integer & Pixel \\
\hline
Polygon metadata & polygons[].action	&Meta data of the polygon action & List & -\\
\hline
Polygon country & polygons[].action.country&	Country of actor of the polygon & String & -\\
\hline
Polygon region & polygons[].action.region&	Region of actor of the polygon & String & -\\
\hline
Polygon date & polygons[].action.date	&Date / time of polygon & String & Date\\
\hline
Polygon author & polygons[].action.actorId &	ID of the actor of the polygon & Integer & -\\
\hline
\end{tabular}
\caption{\label{tab:input.json} Data attributes and description of the {\tt input-google.json} and {\tt input-ign.json} data files.}
\end{table}

\begin{table}[ht]
\centering
\begin{tabular}{|l|l|l|l|l|}
\hline
{\bf Field} & {\bf Attribute name} & {\bf Description} & {\bf Format} & {\bf Unit} \\
\hline
Image ID &id &	ID of the image & String & -\\
\hline
Points list & clicks[]	&List of points whose score is greater than the threshold & List & -\\
\hline
Point Pixel x coordinate & clicks[].x	& x position of point in image & Integer & Pixel\\
\hline
Point Pixel y coordinate & clicks[].y	& y position of point in image & Integer & Pixel\\
\hline
Point score & clicks[].score	&Value of the Pixel annotation consensus for this point. & Float & -\\
&&This value is greater than the chosen threshold &&\\
&&  and lower than the number of clicks on this image.&&\\
\hline
\end{tabular}
\caption{\label{tab:click-analysis.json} Data attributes and description of the {\tt click-analysis.json} and {\tt click-analysis-thres=1.0.json} data files.}
\end{table}

\begin{table}[ht]
\centering
\begin{tabular}{|l|l|l|l|l|}
\hline
{\bf Field} & {\bf Attribute name} & {\bf Description} & {\bf format} & {\bf Unit} \\
\hline
Image ID & id	& ID of the image & String & -\\
\hline
Polygons list & polygons[]	& List of filtered polygons & List & -\\
\hline
Polygon's points list & polygons[].points[]	& List of the points composing the polygon & List & -\\
\hline
Polygon's point pixel x coord. &polygons[].points[].x	&x position of one point of the polygon& Integer & Pixel\\
\hline
Polygon's point pixel y coord.&polygons[].points[].y	&y position of one point of the polygon & Integer & Pixel\\
\hline
Polygon area & polygons[].area	&Area of the polygon & Float & Pixel\\
\hline
Polygon score & polygons[].score	&Value of the Pixel annotation consensus for this & Float & -\\
&&polygon. This value is greater than the chosen the &&\\
&& threshold and lower than the number of clicks &&\\
&& on this image.&&\\
\hline
\end{tabular}
\caption{\label{tab:polygon-analysis.json} Data attributes and description of the {\tt polygon-analysis.json} and {\tt polygon-analysis-thres=1.0.json} data files.}
\end{table}

\begin{table}[ht]
\centering
\begin{tabular}{|l|l|l|l|l|}
\hline
{\bf Field}& {\bf Attribute name} & {\bf Description} & {\bf Format} & {\bf Unit} \\
\hline

Installation ID & idInstallation	&The ID of the installation & Integer & - \\
\hline
Identifier & identifiant	&The name of the image of the installation & String & -\\
\hline
Inverter ID & idInverter	&The ID of the inverter of the installation & Integer & -\\
\hline
Inverter name & nameInverter&	The name of the inverter of the installation & String & -\\
\hline
Number of inverters & countInverters&	The number of inverters attached to the installation & Integer & -\\
\hline
Arrays ID &idArrays	&The ID of the solar arrays used by the installation& Integer & -\\
\hline
Arrays' name &nameArrays	&The name of the solar arrays used by the installation& Float & -\\
\hline
Number of arrays&countArrays	&The number of PV arrays (modules) of the installation& Integer & -\\
\hline
Surface&surface	&The surface (in square meters) of the installation& Float & $m^2$\\
\hline
Azimuth&azimuth	&The azimuth angle in degrees relative to the north &Float&Degrees\\
&&(south = 180) of the installation.& & \\
\hline
Installation type&typeInstallation&	Indicates on which infrastructure the installation & Integer & -\\
&&is mounted:&& \\
&&-	0: rooftop&&\\
&&-	1: unknown&&\\
&&-	2: rooftop of a non-livable building&&\\
&&-	3: ground&&\\
&&-	4: other&&\\
&&-	5: shade house&&\\
&&-	6: sunshade&&\\
&&-	7: solar tracker with 1 axis&&\\
&&-	8: solar tracker with 2 axes&&\\
\hline
Tilt &tilt&	The tilt angle (in degrees) of the installation & Integer & Degrees\\
\hline
Installed capacity &kWp	&The installed capacity of the installation in kWp& Float & kWp\\
\hline
Date of installation & dateInstalled&	The date (month, year) the installation has been installed& String & Date\\
\hline
Is integrated & isIntegrated &	Indicates if the installation is integrated (on the rooftop)& Boolean & -\\
\hline
Self-consumption & selfConsumption &	Indicates if the installation is used for self-consumption & Boolean & -\\
&& (alternative is that PV power is reinjected into the grid) && \\
\hline
{\it Département} & departement &	The {\it département} (county) in which the installation is located& Integer & -\\
\hline
City & city &	The city where the installation is located& String (UTF-8) & -\\
\hline
\end{tabular}
\caption{\label{tab:raw-metadata.csv} Data attributes and description of the {\tt raw-metadata.csv} data file.}
\end{table}

\begin{table}[ht]
\centering
\begin{tabular}{|l|l|l|l|l|}
\hline
{\bf Field}& {\bf Attribute name} & {\bf Description} & {\bf Format} & {\bf Unit} \\
\hline

Installation ID & idInstallation	&The ID of the installation & Integer & - \\
\hline
Identifier & identifiant	&The name of the image of the installation & String & -\\
\hline
Inverter ID & idInverter	&The ID of the inverter of the installation & Integer & -\\
\hline
Inverter name & nameInverter&	The name of the inverter of the installation & String & -\\
\hline
Number of inverters & countInverters&	The number of inverters attached to the installation & Integer & -\\
\hline
Arrays ID &idArrays	&The ID of the solar arrays used by the installation& Integer & -\\
\hline
Arrays' name &nameArrays	&The name of the solar arrays used by the installation& Float & -\\
\hline
Number of arrays&countArrays	&The number of PV arrays (modules) of the installation& Integer & -\\
\hline
Surface&surface	&The surface (in square meters) of the installation& Float & $m^2$\\
\hline
Azimuth&azimuth	&The azimuth angle in degrees relative to the north &Float&Degrees\\
&&(south = 180) of the installation.& & \\
\hline
Installation type&typeInstallation&	Indicates on which infrastructure the installation & Integer & -\\
&&is mounted:&& \\
&&-	0: rooftop&&\\
&&-	1: unknown&&\\
&&-	2: rooftop of a non-livable building&&\\
&&-	3: ground&&\\
&&-	4: other&&\\
&&-	5: shade house&&\\
&&-	6: sunshade&&\\
&&-	7: solar tracker with 1 axis&&\\
&&-	8: solar tracker with 2 axes&&\\
\hline
Tilt &tilt&	The tilt angle (in degrees) of the installation & Integer & Degrees\\
\hline
Installed capacity &kWp	&The installed capacity of the installation in kWp& Float & kWp\\
\hline
Date of installation & dateInstalled&	The date (month, year) the installation has been installed& String & Date\\
\hline
Is integrated & isIntegrated &	Indicates if the installation is integrated (on the rooftop)& Boolean & -\\
\hline
Self-consumption & selfConsumption &	Indicates if the installation is used for self-consumption & Boolean & -\\
&& (alternative is that PV power is reinjected into the grid) && \\
\hline
{\it Département} & departement &	The {\it département} (county) in which the installation is& Integer & -\\
&& located && \\
\hline
City & city &	The city where the installation is located& String (UTF-8) & -\\
\hline
Controlled&Controlled	& Indicates whether the installations' metadata are clean & Boolean & -\\
\hline
Matched with IGN &IGNControlled	&Indicates whether the installation corresponds to a unique & Boolean & - \\
image&& segmentation mask corresponding to an IGN image&&\\
\hline
Matched with Google &GoogleControlled	&Indicates whether the installation corresponds to a unique & Boolean & -\\
image&& segmentation mask corresponding to a Google image&&\\
\hline
\end{tabular}
\caption{\label{tab:metadata.csv} Data attributes and description of the {\tt metadata.csv} data file.}
\end{table}

\end{document}